\documentclass[acmtog,screen]{acmart}
\acmSubmissionID{1086}

\usepackage{booktabs} % For formal tables

\usepackage{enumitem}
\usepackage{amsmath}
\usepackage{amsfonts}
\usepackage{multirow}
\usepackage{bigdelim}
\usepackage{wrapfig}
\usepackage{xspace}
\usepackage{color}
\usepackage{booktabs}
\usepackage{dsfont}
\usepackage{ifthen}
\usepackage{subcaption}
\usepackage{bbold}
\usepackage{xcolor, pgf}
\usepackage{color, colortbl}
\usepackage[normalem]{ulem} % for sout
\usepackage{placeins}
\usepackage{soul}
\usepackage[many]{tcolorbox}
\usepackage[noabbrev,capitalise,nameinlink]{cleveref}
\creflabelformat{equation}{#2\textup{#1}#3}% 
\Crefname{lstlisting}{Listing}{Listings}

\usepackage{titlesec}
\usepackage{adjustbox}
\usepackage{arydshln}

\usepackage{algorithm}
\usepackage{algorithmicx}
\usepackage{algpseudocode}
\algnewcommand\algorithmicinput{\textbf{Input:}}
\algnewcommand\INPUT{\item[\algorithmicinput]}
\algnewcommand\algorithmicoutput{\textbf{Output:}}
\algnewcommand\OUTPUT{\item[\algorithmicoutput]}
\algnewcommand\algorithmicforeach{\textbf{for each}}
\algtext*{EndFor}% Remove "end for" text

\algdef{S}[FOR]{ForEach}[1]{\algorithmicforeach\ #1\ \algorithmicdo}
\algrenewcommand{\alglinenumber}[1]{\color{red!80!blue}\footnotesize#1:}

\algnewcommand\Func[2]{\textcolor{green}{#1}\textcolor{green}{(#2)}}
\algnewcommand\Insert[2]{Insert {#1} to #2.}
\algnewcommand\Input[1]{\State \textbf{Input: } #1}
\algnewcommand\Output[1]{\State \textbf{Output: } #1}

\newlength\myboxwidth
\definecolor{gray}{rgb}{0.5,0.5,0.5}
\definecolor{green}{rgb}{0, 0.6, 0}
\definecolor{orange}{rgb}{1, 0.5, 0}
\definecolor{mahogany}{rgb}{0.75, 0.25, 0.0}
\definecolor{purple}{rgb}{0.6, 0, 0.6}
\definecolor{darkgreen}{rgb}{0, 0.3, 0}
\definecolor{orange}{rgb}{1, 0.5, 0.}
\definecolor{lightblue}{rgb}{0.52, 0.75,0.91}
\definecolor{secondblue}{rgb}{0.52, 0.78,0.70}
\definecolor{softgreen}{rgb}{0.66,0.87,0.74}
\definecolor{softred}{rgb}{0.96,0.71,0.69}

\colorlet{soullightblue}{lightblue!50}
\newcommand{\besthint}[1]{\sethlcolor{soullightblue}\hl{#1}}
\colorlet{soulsecondblue}{secondblue!50}
\newcommand{\sechint}[1]{\sethlcolor{soulsecondblue}\hl{#1}}

\newcommand{\bestcell}[1]{\cellcolor{lightblue!50}#1}
\newcommand{\seccell}[1]{\cellcolor{secondblue!50}#1}

\definecolor{purpleD}{rgb}{.8941, .8, .9043}
\definecolor{greenD}{rgb}{.7145, .9249, .7999}

\newcommand{\ignore}[1]{}
\newcommand{\none}[1]{}
\newcommand{\com}[1]{}
\newcommand{\etal}{{\it{et~al.}}}
\newcommand{\ie}{i.e.,}
\newcommand{\eg}{e.g.,}

\newcommand{\inputprompt}{\mathcal{P}_{\mathrm{input}}}

\newcommand{\completesketch}{\mathcal{S}_{\mathrm{complete}}}

\newcommand{\inputsketch}{\mathcal{S}_{\mathrm{input}}}
\newcommand{\optsketch}{\mathcal{S}_{\mathrm{opt}}}
\newcommand{\guideimg}{\mathcal{I}_{\mathrm{guide}}}
\newcommand{\rastersketch}{\mathcal{I}_\mathrm{sketch}}

\colorlet{soullightyellow}{yellow!50}
\newcommand{\asiahl}[1]{#1}
\newcommand{\rvhl}[1]{#1}
\newcommand{\arik}[1]{{\color{orange}#1}}

\renewcommand{\arik}[1]{#1}
\tcbset{top=0mm,bottom=0mm,left=0mm,right=0mm,fonttitle=\bfseries\scriptsize\color{gray},colbacktitle=white,enhanced,attach boxed title to top center={yshift=-1mm},boxed title style={top=0mm,bottom=0mm,left=0mm,right=0mm}}

\usepackage{listings}
\definecolor{codegreen}{rgb}{0,0.6,0}
\definecolor{codegray}{rgb}{0.5,0.5,0.5}
\definecolor{codepurple}{rgb}{0.58,0,0.82}
\definecolor{backcolour}{rgb}{0.95,0.95,0.92}

\usepackage{lstautogobble}  % Fix relative indenting
\usepackage{color}          % Code coloring
\usepackage{zi4}            % Nice font

\definecolor{bluekeywords}{rgb}{0.13, 0.13, 1}
\definecolor{greencomments}{rgb}{0, 0.5, 0}
\definecolor{redstrings}{rgb}{0.9, 0, 0}
\definecolor{graynumbers}{rgb}{0.5, 0.5, 0.5}

\lstdefinestyle{mystyle}{
    backgroundcolor=\color{backcolour},   
    commentstyle=\color{codegreen},
    keywordstyle=\color{magenta},
    numberstyle=\tiny\color{codegray},
    stringstyle=\color{codepurple},
    basicstyle=\ttfamily\footnotesize,
    breakatwhitespace=false,         
    breaklines=true,                 
    captionpos=b,                    
    keepspaces=true,                 
    numbers=left,                    
    numbersep=5pt,                  
    showspaces=false,                
    showstringspaces=false,
    showtabs=false,                  
    tabsize=2
}

\lstdefinestyle{anotherstyle}{
    autogobble,
    columns=fullflexible,
    showspaces=false,
    showtabs=false,
    breaklines=true,
    showstringspaces=false,
    breakatwhitespace=true,
    escapeinside={(*@}{@*)},
    commentstyle=\color{greencomments},
    keywordstyle=\color{bluekeywords},
    stringstyle=\color{redstrings},
    numberstyle=\color{graynumbers},
    basicstyle=\ttfamily\footnotesize,
    frame=l,
    framesep=12pt,
    xleftmargin=12pt,
    tabsize=4,
    captionpos=b
}

\newcommand{\methodName}{AutoSketch\xspace}

\usepackage[frozencache,cachedir=.]{minted}
\usepackage{etoolbox}
\newminted{python}{autogobble,escapeinside=||,fontsize=\footnotesize}
\BeforeBeginEnvironment{minted}{
    \begin{tcolorbox}[arc=1pt,colback=black!3!white,colframe=black!16!white,boxrule=0.3mm,before skip=1mm,after skip=1mm,top=.5mm,bottom=.5mm,left=1mm,right=1mm]
}
\AfterEndEnvironment{minted}{
    \end{tcolorbox}
}
\acmJournal{TOG}
\acmVolume{38}
\acmNumber{4}
\acmArticle{39}
\acmYear{2021}
\acmMonth{7}

\setcopyright{none}

\begin{document}

\title[\methodName: VLM-assisted Style-Aware Vector Sketch Completion]{\methodName: VLM-assisted Style-Aware Vector Sketch Completion}

\author{Hsiao-Yuan Chin}
\authornote{Both authors contributed equally to this research.}
\orcid{0009-0004-7313-5528}
\affiliation{%
 \institution{National Taiwan University}
 \country{Taiwan}
}
\author{I-Chao Shen}
\authornotemark[1]
% \authornote{Corresponding author.}
\orcid{0000-0003-4201-3793}
\affiliation{%
 \institution{The University of Tokyo}
 \country{Japan}
}
\author{Yi-Ting Chiu}
\orcid{0009-0004-0937-2705}
\affiliation{%
 \institution{National Taiwan University}
 \country{Taiwan}
}
\author{Ariel Shamir}
\orcid{0000-0003-4201-3793}
\affiliation{%
 \institution{Reichman University}
 \country{Israel}
}
\author{Bing-Yu Chen}
\orcid{0000-0003-0169-7682}
\affiliation{%
 \institution{National Taiwan University}
 \country{Taiwan}
}

\begin{abstract}
\arik{Sketches are an important medium of expression and recently many works concentrate on automatic sketch creations.}
\arik{One such ability very useful for amateurs is text-based completion of a partial sketch to create} a complex scene, while \asiahl{preserving the style of the partial sketch}.
Existing methods \asiahl{focus solely on generating sketch that match the content in the input prompt in a predefined style}, \asiahl{ignoring the styles of the input partial sketches, \eg~the global abstraction level and local stroke styles.}
To address this challenge, we introduce \methodName, a style-aware vector sketch completion method that accommodates diverse sketch styles and supports iterative sketch completion.
\methodName \asiahl{completes the input sketch in a style-consistent manner using a two-stage method.} 
\asiahl{In the first stage}, we initially optimize the strokes to match an input prompt augmented by style descriptions extracted from a vision-language model (VLM).
\asiahl{Such style descriptions lead to non-photorealistic guidance images which enable more content to be depicted through new strokes.}
\arik{In the second stage}, we utilize the VLM to adjust the strokes from the previous stage to adhere to the style \asiahl{present in the input partial sketch through an iterative style adjustment process.
In each iteration, the VLM identifies a list of style differences between the input sketch and the strokes generated in the previous stage, translating these differences into adjustment codes to modify the strokes.}
We compare our method with existing methods using various sketch styles and prompts, perform extensive ablation studies and qualitative and quantitative evaluations, and demonstrate that \methodName can support \arik{diverse} sketching scenarios.
\end{abstract}

%
% End generated code
%


%
% Uncomment the following part to add your teaser.
%
\begin{teaserfigure}
  \includegraphics[width=\textwidth]{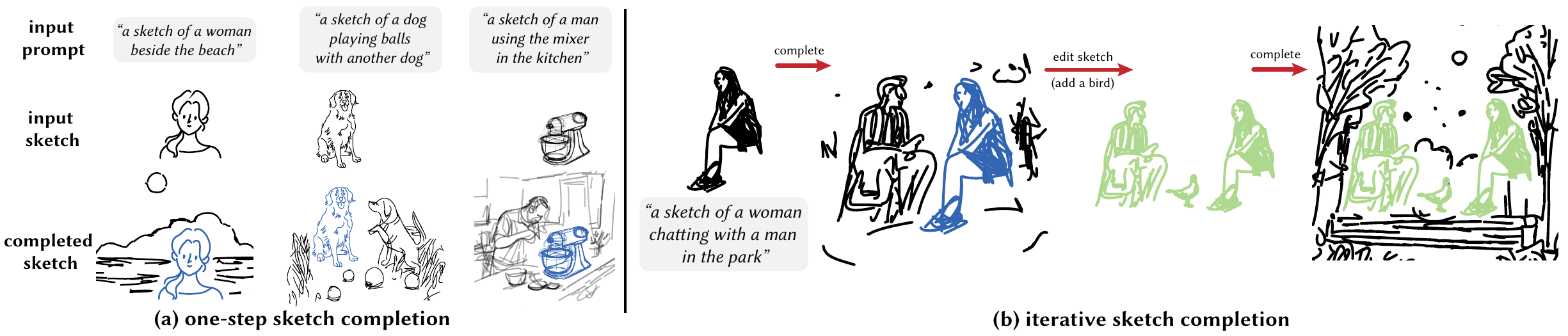}
  \caption{
    (a) Given an input prompt and a sketch, our method completes the input sketch by accurately representing the input prompt and maintain the style of the input partial sketch.
    (b) Users iteratively employ \methodName to create a complex sketch. 
    For example, after the first completed sketch is generated, the user decide to retain the strokes representing the man and woman, draw a bird, and our method completes the sketch by adding strokes depicting the trees and grass.
    (The blue and green strokes denote the first and second iterations of the input sketches.)
  }
  \label{fig:teaser}
\end{teaserfigure}

\maketitle

%!TEX root = paper.tex
\section{Introduction}
\label{sec:intro}
Sketching has long been a key form of visual expression that rapidly communicates ideas and expresses concepts.
Even people with little experience can easily sketch simple objects and ideas.
However, creating sketches that depict complex scenes remains a significant challenge for many.
Typically, individuals begin sketching by creating a rough partial sketch but often struggle to turn this into a final complex sketch that maintains a unique style.

Recent text-based sketch generation methods~\cite{xing2023diffsketcher, qu2023sketchdreamer, jain2023vectorfusion} \asiahl{leverage user-provided text prompts to generate intricate sketches either from scratch or progressively.}
However, these methods do not adequately consider input partial sketches, thus creating two major issues.
First, they often generate redundant strokes that duplicate elements already present in the input partial sketch.
Second, they ignores the styles of the input sketch, \eg~\rvhl{the global level of abstraction and the local stroke styles, such as stroke thickness, smoothness, curvature, opacity.}
% stroke geometric- and appearance-properties. 
Therefore, the styles of the generated strokes do not align with those of the input partial sketch.

To address these issues, we propose \methodName, a novel style-aware vector sketch completion method that takes a text prompt and a partial sketch as input.
Our method completes the partial sketch by generating strokes that illustrate missing elements or concepts \arik{according to the prompt}, while preventing the creation of redundant strokes and ensuring that the style aligns with that of the input sketch.
% \rvhl{In our work, we focus on the the global levels of abstraction and local stroke styles, such as stroke thickness, smoothness, curvature, opacity.}

We observed that the stroke optimization method using fixed text augmentation adopted by~\cite{xing2023diffsketcher, qu2023sketchdreamer, jain2023vectorfusion} \asiahl{often results in unsatisfied completed sketches.}
\asiahl{In terms of content, the generated guidance image often contains blurred area or lacks clear boundaries, leading to some elements being overlooked during the sketch generation process.
In terms of style, the use of a single type of stroke parameterization causes} the style of the generated strokes to misalign with those of the input sketch.

Based on this observations, we utilize a pretrained vision-language model (VLM) along with a stroke optimization process.
In the first stage, we augment the input prompt with VLM-generated style descriptions based on the input partial sketch.
This augmentation allows ControlNet~\cite{zhang2023adding} to generate guidance images that \asiahl{exhibit some non-photorealistic rendering styles} for stroke generation.
Next, we optimize the strokes based on these guidance images.
\asiahl{To resolve the redundant strokes issue,} we introduce an \asiahl{overlap} penalty to ensure that the generated strokes do not overlap with those of the input partial sketch.
In the second stage, we employ the VLM to iteratively adjust the styles of the optimized sketch. 
\asiahl{We task the VLM with identifying style differences between the input sketch and the most recent optimized sketch}.
Then, we ask the VLM to generate adjustment code based on the identified style differences and apply them to the latset optimized sketch to obtain the final sketch.
\asiahl{While the VLM typically adjusts styles effectively in one iteration, this iterative process ensures effective style adjustment due to the inherent instability of the VLM.}

\asiahl{The advantage of using a VLM for style adjustment, compared to existing optimization-based methods, is its ability to support both continuous and discrete adjustments, including stroke deletion or splitting.}
\asiahl{Although prior work~\mbox{\cite{vinker2024sketchagent,cai2023leveraging}} has shown that it is technically feasible to prompt a pretrained VLM to generate adjusted strokes directly, these methods are constrained by token limitations, which prevents them from generating complex sketches with a large number of strokes.
}
 By generating \emph{style adjustment code} with the VLM, we can overcome these challenges, resulting in a more stylistically consistent sketch without losing essential content.

We compare our results with those of existing methods across various sketch styles and prompts.
Extensive quantitative and qualitative evaluations reveal that the completed sketches generated by our method better preserve the style of the input partial sketches and more accurately represent the contents specified by the prompts.

\section{Related Work}
\label{sec:related}
\subsection{Vector Sketch Generation}
Previous studies~\cite{eitz2012hdhso,ha2018neural,sketchy2016} have collected sketch datasets of amateur sketches that sought to realistically depict everyday objects, while OpenSketch~\cite{gryaditskaya2019opensketch} contains professional sketches of product designs.
Existing studies used these sketch datasets and various deep learning models~\cite{ha2018neural,lin2020sketch,ribeiro2020sketchformer,zhou2018learning} to generate sketch sequences.
However, given their reliance on these sketch datasets, such methods generally generate sketches of only simple objects.

Recently, novel methods~\cite{vinker2022clipasso,vinker2023clipascene, xing2023diffsketcher,sketchVideo24,frans2022clipdraw,qu2023sketchdreamer} employs the ``synthesis-through-optimization'' paradigm have emerged.
These methods typically optimize stroke geometry and appearance using priors derived from large pretrained models such as CLIP~\cite{radford2021learning}, and text-to-image~\cite{rombach2022high} and text-to-video~\cite{wang2023modelscope} models.
\rvhl{However, these methods either create sketches from scratch to fit the input prompt or modify sketches based on the updated prompt~\mbox{\cite{mo2024text}} in a predefined style, while neglecting the styles present in the input sketches.}
% However, these methods primarily focus on generating \rvhl{or editing} sketches that match the content
% of the input prompt in a predefined style, while neglecting the styles present in the input sketches.

\subsection{Sketch Styles}
In the past works, style is often discussed in terms of two components: local curve level and global abstraction level.
At the local curve level, style is characterized by a combination of geometry (shape) and appearance (\eg~strokes and textures).
For example, Li~\etal~\shortcite{li2013curve} identified geometric curve styles in a set of shapes, while Berger~\etal~\shortcite{berger2013style} analyzed both the geometric and appearance sketch styles, as well as global abstraction level for a specific artist in portrait sketching.
Recently, Vinker~\etal~\shortcite{vinker2023clipascene} introduced a method for generating scene sketches that vary in levels of abstraction.
Our method considers both the global abstraction level and the local geometric and appearance styles of sketches, ensuring that the style aligns between the input sketch and generated sketch.

\subsection{LLM-based Sketch and SVG Editing}
Recent advancements in large language models (LLMs) have enabled extensive research on vector graphic generation and editing~\cite{nishina2024svgeditbench,zou2024vgbench,cai2023leveraging,wu2024chat2svg}.
This progress has led to the development of new benchmarks and frameworks aimed at evaluating enhancing the capabilities of LLMs.
For example, StarVector~\cite{rodriguez2025starvector} presents a multimodal LLM designed to vectorize raster images.
Other previous works~\cite{wu2023iconshop,tang2024strokenuwa,xing2024empowering} incorporate specialized tokenization methods or modular architectures to improve LLMs’ understanding of SVG structures, enabling advanced tasks such as text-guided icon synthesis and SVG manipulation. 
However, many of these methods rely on additional large-scale training data to finetuning LLMs.
In contrast, SketchAgent~\cite{vinker2024sketchagent} and Chat2SVG~\cite{wu2024chat2svg} use off-the-shelf LLMs without finetuning but is limited to simple concepts with a small number of strokes.
Our method, without any finetuning, can handle partial sketches involving a larger number of strokes, enabling complex scenes with complex object interactions and compositions.

% \subsection{Style Transfer and Aesthetic Consistency}
% \ichao{Maybe our method can be seen as doing style transfer using code...}
% Visual style transfer traditionally focused on raster images, using convolutional neural networks to blend the style of a reference painting with the content of a target image~\cite{gatys2016image}. Recent efforts have also explored style adaptation in the vector domain. For instance, stroke-based style transfer methods have been introduced to manipulate line quality and stylistic attributes in vector line art~\cite{sonobe2022stroke}, though many approaches still rely on fixed style exemplars or manual adjustments. Previous vector sketch methods did not explicitly address style continuity—particularly when dealing with partially complete sketches. Our approach incorporates a novel style loss derived from VLM embeddings, enabling us to measure and preserve stylistic similarity at a higher-level semantic feature space rather than relying solely on low-level pixel intensities. This higher-level approach promotes consistency in line quality, stroke thickness, and thematic coherence, ensuring that newly generated elements appear as a natural extension of the user’s partial input.
\begin{figure*}[!h]
  \centering
  \includegraphics[width=\linewidth]{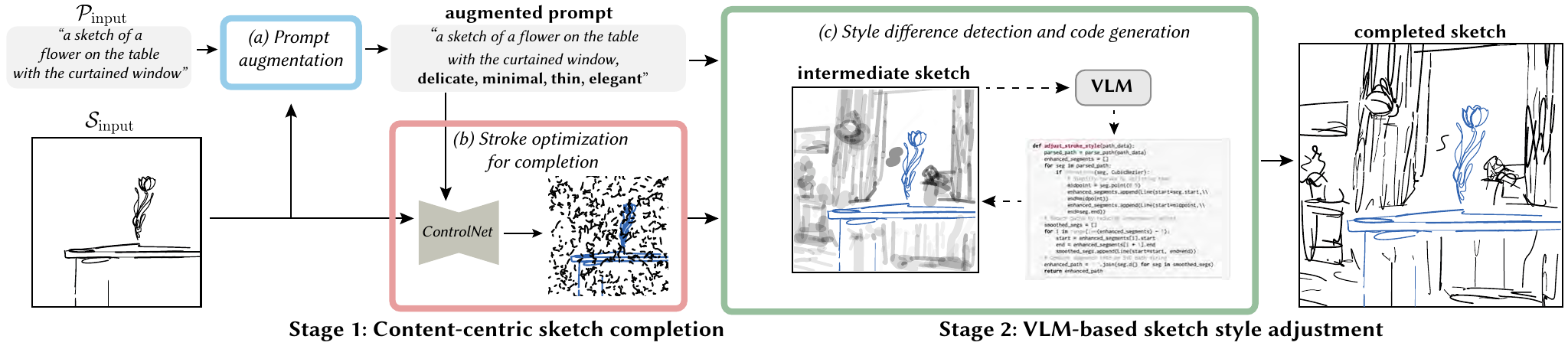}
  \caption{
\textbf{Overview of our method.}
Given a user-provided prompt $\inputprompt$ and a partial sketch $\inputsketch$, our method first (a) stylizes the input prompt by augmenting it using style descriptions generated by the VLM (\textbf{bold text}).
Using the augmented prompt, the method then performs (b) stroke optimization to generate strokes that fill the missing regions, thus ensuring that the intermediate sketch can fully represents the content of the input prompt.
To align the styles of the intermediate sketch and the input sketch, we (c) iteratively instruct the VLM to \asiahl{identify style differences and translate them into} style adjustment codes that modifies the strokes of the intermediate sketch.
Finally, we obtain a final completed sketch.
}
\label{fig:overview}
\end{figure*}

% \subsection{Summary of the Pipeline}
% \label{sec:method_summary}

% Our pipeline thus consists of three stages:
% \begin{enumerate}
%     \item \textbf{Text and Image Guidance Generation:} 
%     Transforms the partial sketch into a stylized text prompt via the VLM and obtains an image guidance $\mathcal{I}_g$ through a conditional text-to-image generator.
%     \item \textbf{Content Completion:} 
%     Uses a differentiable renderer to optimize newly added strokes under a composite objective function that includes CLIP-based semantic alignment, perceptual matching, and a mask penalty.
%     \item \textbf{Style-Aware Adjustment:} 
%     Employs the VLM with a system prompt to refine the final sketch, ensuring consistency with the style of the original partial sketch.
% \end{enumerate}

\section{Overview}
In~\autoref{fig:overview}, we illustrate the overview of our method.
Our method takes as input a text prompt $\inputprompt$ and a partial sketch $\inputsketch$.
The text prompt describes the content to be illustrated in the completed sketch, and the input partial sketch represents only partial content described in the prompt.
The output is a completed sketch $\completesketch=\inputsketch \cup \optsketch$ that fully represents the content of $\inputprompt$ in a coherent style.
Our method has two stages: \textit{\asiahl{content-centric} sketch completion} and \textit{sketch style adjustment}.

 % Our goal is to optimize a set of parametric strokes to automatically generate a vector
% sketch that matches the description of the text prompt.
% proceeds in three main stages to achieve style-aware vector sketch completion. 
In the first stage, the goal is to optimize a set of parametric strokes that, when combined with the input partial sketch, ensure that the complete sketch represents the content of $\inputprompt$ without considering the sketch styles.
% We utilize a pre-trained text-to-image diffusion model as the diffusion prior to 
First, we augment $\inputprompt$ by leveraging a large vision-language model (VLM) to produce style descriptions of the input partial sketch $\inputsketch$ (\autoref{fig:overview}(a)).
% \ie~$\finalprompt=\{\inputprompt \cup \augprompt\}$ (\autoref{fig:overview}(a)).
% This step extracts style-related cues and integrates them into a stylized text prompt.
Then, we optimize the parameters of \asiahl{a set of randomly sampled strokes} using a diffusion prior conditioned on the augmented text prompt (\autoref{fig:overview}(b)) and obtains an intermediate sketch. %$\intermediatesketch$.

In the second stage, the goal is to adjust the styles of the intermediate sketch to achieve a cohesive look throughout the completed sketch. 
We task the VLM to perform a style adjustment on the intermediate sketch.
The VLM \asiahl{begins with identifying the style differences between strokes in the input sketch and the optimized strokes.}
The VLM then generates an adjustment code based on \asiahl{the detected style differences}.
We then apply the adjustment code to the intermediate sketch.
\asiahl{
We instruct the VLM to focus on differences in global abstraction levels and local stroke styles. 
In most of the situation, the VLM effectively identifies style differences and translates them into adjustment codes. 
However, it occasionally overlooks some differences when creating these codes.
To resolve this issue, we repeat this process until the sketches are no longer updated.
}
% \section{Method}
% \label{sec:method}

\section{Stage 1: Content-centric Sketch Completion}
Inspired by previous works~\cite{xing2023diffsketcher,vinker2023clipascene}, we optimize the parameters of a group of strokes by leveraging the prior of a pretrained text-to-image (T2I) model.
Unlike previous works, our method employs a user-provided partial sketch $\mathcal{S}_{\text{input}}$ as an additional input.
Therefore, we employ a conditional T2I model (\eg~ControlNet Scribble\footnote{\url{https://huggingface.co/xinsir/controlnet-scribble-sdxl-1.0}}) to optimize the stroke parameters.

\subsection{Prompt Augmentation}
\label{sec:method_step1}
Although the conditional T2I model generates images that match the input text prompt $\inputprompt$, we observed that these images are often unsuitable for directly generating strokes.
The main reason is that the T2I model generates images in a photorealistic style, which tends to include many blurs and lacks clear boundaries.
Previous SDS-based methods~\cite{qu2023sketchdreamer, jain2023vectorfusion} have attempted to enhance sketch generation by augmenting the input prompt with a fixed term to promote a more sketch-like style.
Such augmentation works effectively because they focus on a single type of sketch style. 
\asiahl{However, since our method needs to accommodate input sketches with various styles, using a fixed augmentation to promote non-photorealistic styles, is not effective.
}
To address this issue, we augment the input prompt $\mathcal{P}_{\text{input}}$ with some style descriptions using the VLM \asiahl{based on the input partial sketch} (\autoref{fig:overview}(a)).
Specifically, we render the input partial sketch $\inputsketch$ into a raster image and then ask the VLM to generate textual descriptions capturing \asiahl{specifically the global level of abstraction} and local stylistic cues of the rendered image.
Then, we augment these style descriptions to the input prompt to obtain the augmented prompt.
% We use $\finalprompt$ to denote the final prompt in the following paragraph.
% The final prompt becomes: \ie~$\finalprompt=\{\mathcal{P}_{\text{input}} \cup \mathcal{P}_{\text{aug}}\}$.
% \todo{Maybe we can add a figure contains some examples here?}
% In~\cref{?}, we show several examples of extracted style descriptions using VLM.

% \paragraph{Image Guidance via Conditional Generation.}
% \ichao{Merge this part with next paragraph.}
% Next, we employ a conditional text-to-image generation model (\eg~ControlNet Scribble) to produce an image guidance $\mathcal{I}_g$:
% \begin{align}
%     \mathcal{I}_g = G(\mathcal{S}, \mathcal{T}),
% \end{align}
% where $G(\cdot)$ denotes the conditional generator that takes the partial sketch $\mathcal{S}$ (e.g., as a scribble or mask) and the stylized text prompt $\mathcal{T}$ as inputs. The resulting image $\mathcal{I}_g$ encodes shape, layout, and stylistic attributes consistent with both the original partial sketch and the textual description. This image guidance will serve as a visual reference during content completion.

\begin{figure}[!t]
  \centering
  \includegraphics[width=\linewidth]{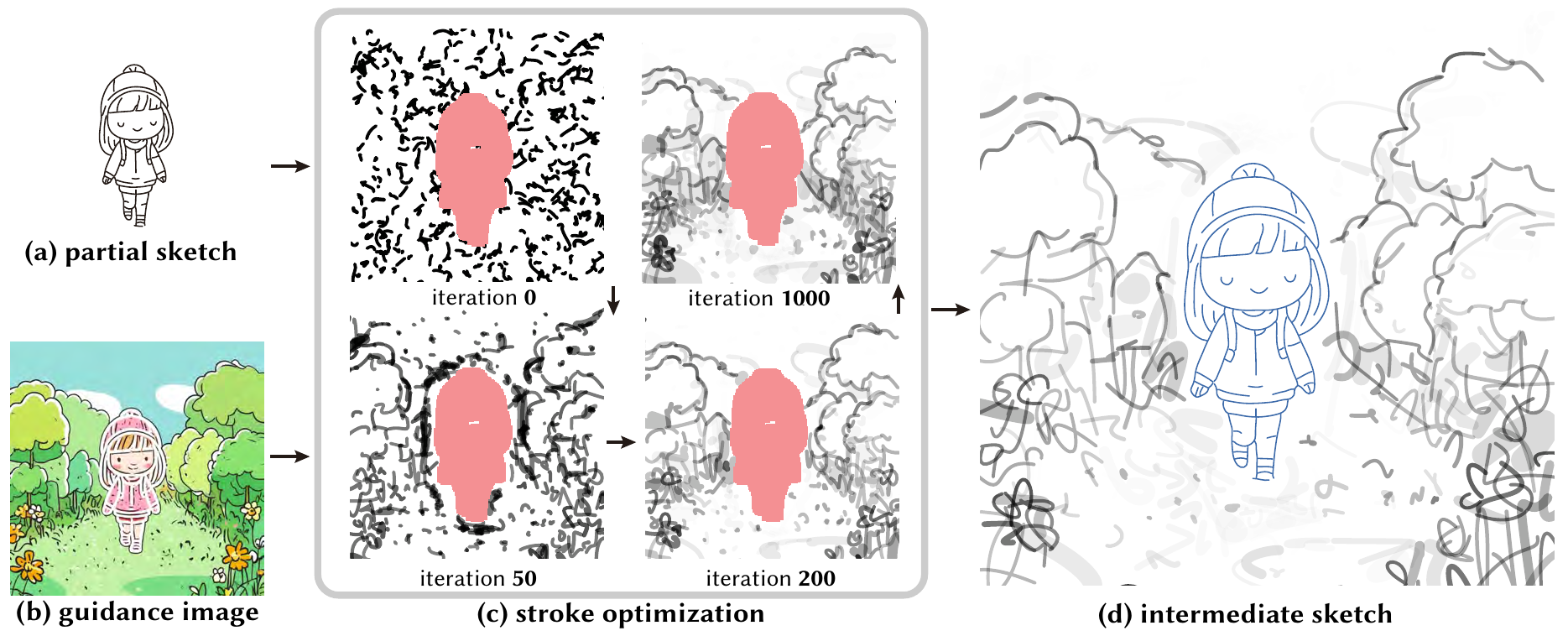}
  \caption{
  % \todo{Maybe we can remove this one..}
\textbf{Overview of stroke optimization.} 
Given (a) the input partial sketch and (b) the generated guidance image, our method (c) iteratively updates the position, opacity, and width of each stroke.
This ensures that (d) the resulting intermediate sketch aligns with the guidance image visually but does not overlap with the input partial sketch. 
}
  \label{fig:stroke_opt}
\end{figure}
\subsection{Stroke Optimization for Completion}
\label{sec:method_step2}
Using the augmented prompt, we generate strokes that fill the empty regions of the input partial sketch.
We define the \asiahl{set of $n$} strokes to be optimized as $\optsketch=\{s_1,\dots,s_n\}$, and each stroke is defined as:
\begin{align}
    s_i = \left\{ \{p^j_i\}_{j=1}^4,o_i, w_i \right\},
\end{align}
where $\{p^j_i\}_{j=1}^4$ are the control points of a cubic Bézier curve, $o_i$ denotes an opacity attribute, and $w_i$ denotes the stroke width.
Initially, we generate a guidance image $\guideimg$ using a conditional T2I model based on the augmented prompt.
Then, we optimize all parameters of $\optsketch$ to obtain a sketch that is consistent with the guidance image $\guideimg$ (\autoref{fig:stroke_opt}).
\asiahl{For initialization, we randomly place all strokes on the canvas.}
At iteration $t$, we rasterize the strokes using a differentiable rasterizer $R$ to generate the raster sketch: $\rastersketch = R(\completesketch)$, and we optimize the following objective function when updating the strokes:
\begin{align}
L_{\mathrm{all}} &= \alpha(1 - \text{sim}\bigl(\phi_{\mathrm{vis}}(\rastersketch), \phi_{\mathrm{vis}}(\guideimg)\bigr)) \nonumber \\ 
&+ \beta(LPIPS(\rastersketch, \guideimg)) + \gamma \sum_{x_k \in \mathbf{x}} \mathds{1} \left[ \mathbf{M}(x_k) = 1 \right]
\label{eq:complete_func}
\end{align}
% \begin{align}
%     L_{\mathrm{all}}
%     &= \alpha \;\tcboxmath[colback=white,colframe=purpleD,title=CLIP visual alignment]{(1 - \text{sim}\bigl(\phi_{\mathrm{vis}}(\rastersketch), \phi_{\mathrm{vis}}(\guideimg)\bigr))} \label{eq:clip_alignment}\\ 
%     &+ \beta \;\tcboxmath[colback=white,colframe=greenD,title=perceptual loss]{(LPIPS(\rastersketch, \guideimg))} \label{eq:percep}\\ 
%     &+ \gamma \;\tcboxmath[colback=white,colframe=lightblue,title=Overlap penalty]{\sum_{x_k \in \mathbf{x}} \mathds{1} \left[ \mathbf{M}(x_k) = 1 \right]},
%     \label{eq:overlap_penalty}
% \end{align}
% \begin{align}
%     L_{\mathrm{total}}
%     = \alpha \,L_{\mathrm{VCLIP}}
%     + \beta \,L_{\mathrm{perc}}
%     + \gamma \,L_{\mathrm{mask}},
% \end{align}
% \todo{Check which clip is this? image or text?}
where $\alpha,\beta,\gamma$ control the relative importance of the three terms.
The first term measures the visual alignment between the guidance image $\guideimg$ and the raster sketch $\rastersketch$ using the CLIP visual encoder $\phi_{\mathrm{img}}(\cdot)$, where $\text{sim}(\mathbf{x},\mathbf{y})=\frac{\mathbf{x}\cdot\mathbf{y}}{\|\mathbf{x}\|\cdot \|\mathbf{y}\|}$ is the cosine similarity.
Additionally, we further minimize the LPIPS loss to enhance the visual similarity of $\rastersketch$ and $\guideimg$.

% \begin{wrapfigure}{r}{0.10\textwidth}
% \includegraphics[width=0.08\textwidth]{figs/mask_penalty.pdf}
% \end{wrapfigure} 
To ensure that the strokes do not overlap with those of the input partial sketch $\inputsketch$, we introduce an overlap penalty loss. 
Specifically, we first define a binary mask $\mathbf{M}$ that encodes the regions in $\inputsketch$ where strokes already exist:
% and should thus not be altered:
\begin{align}
    \mathbf{M}(x) = 
    \begin{cases}
    1, & \text{if pixel } x \text{ belongs to strokes in } \inputsketch, \\
    0, & \text{otherwise}.
    \end{cases}
    \label{eq:inside_check}
\end{align}

Then, we sample $10$ points on each stroke $s_i \in \optsketch$.
% , we  on it and obtain a set of sample point $\mathbf{x}$.
For each sample point $x_k$, if that point falls in $M$ (the filled black circles in the inset), we introduce a penalty, where $\mathds{1}[\cdot]$ in~\autoref{eq:complete_func} is the indicator function.

\asiahl{
Our objective function $L_{\mathrm{all}}$ is very similar to the one used in DiffSketcher~\mbox{\cite{xing2023diffsketcher}}, with the exception that we have discarded the augmentation SDS (ASDS) loss.
The reason for this is that we found ASDS loss contributed very little to the optimization, as indicated by the results shown in DiffSketcher paper and their released code.
Therefore, to reduce computation time and focus on achieving better sketch completion result, we opted to use the overlap penalty loss instead of the ASDS loss.
}

After optimizing $L_{\mathrm{all}}$, we obtain the intermediate sketch by combining the optimized strokes with those of input partial sketch.
The strokes in the intermediate sketch contain the overall content described in the input prompt, but the styles are not coherent yet.

\section{VLM-based Sketch Style Adjustment}
\label{sec:method_step3}
After optimizing the strokes in~\autoref{sec:method_step2}, we have filled in the empty areas.
However, this does not guarantee that the strokes in the intermediate sketch will exhibit global stylistic coherence.
The wide variety of sketch styles complicates the process of defining appropriate parameterizations that can capture all potential styles.
\asiahl{Moreover, stroke optimization can only adjust stroke parameters continuously and cannot accommodate discrete style changes, such as stroke deletion or simplification.}
To address these challenges, we instruct the VLM to guide the style adjustment of the intermediate sketch.
% Therefore, we use the VLM to adjust the style of $\intermediatesketch$ to align with that of $\inputsketch$.
Specifically, we represent the intermediate sketch in SVG format and request the VLM to modify the SVG to achieve the desired style adjustments.
However, several challenges arise due to the limitations of existing VLMs.
First, existing VLMs can handle only a limited number of tokens, restricting the number of strokes that can be included in the intermediate sketch.
Second, these VLMs often hallucinate, \ie~they may generate strokes absent in the intermediate sketch which does not match the input prompt.

\begin{figure}[!t]
  \centering
  \includegraphics[width=\linewidth]{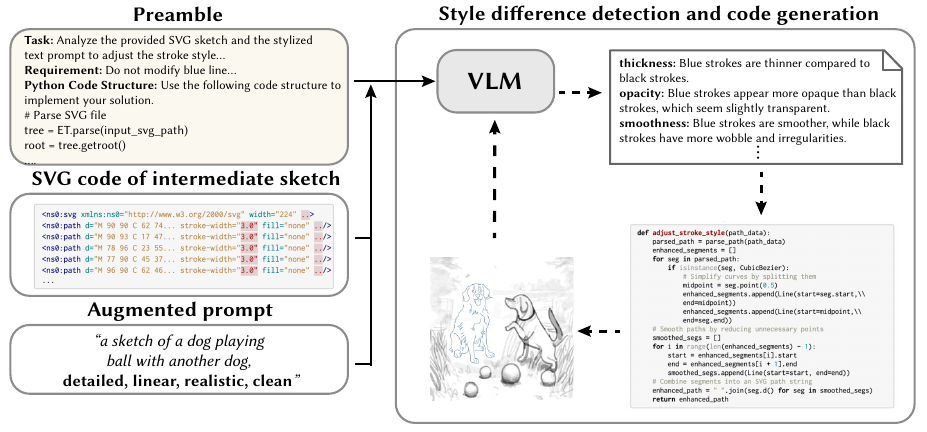}
  \caption{
\textbf{Overview of VLM-based sketch style adjustment.} 
The complete system prompt we provided to the VLM consists of a preamble, an augmented prompt, and the SVG code of the intermediate sketch.
We input this information into the VLM, which then generates detected style differences and adjustment codes to iteratively modify the styles of the sketch.
}
  \label{fig:vlm_adjust}
\end{figure}
To address these issues, \asiahl{we propose an iterative style differences detection and code generation process (\mbox{\autoref{fig:vlm_adjust}}).}
\asiahl{In each iteration, we ask the VLM to generate a list of style differences between the strokes from the previously optimized sketch with those in the input sketch. }
\asiahl{Next, we instruct the VLM to generate style adjustment codes based on the identified style differences.}
Then, we apply the adjustment codes to the previously optimized sketch to obtain the final sketch.
\asiahl{This process will be terminated until sketches are no longer updated.}
% The main purpose of this iterative process is to ensure the styles are adjusted effectively, given the instability of the VLM.}

\paragraph{Style difference detection.}
In this step, we provide the VLM with the following information:
\begin{itemize}
    \item A preamble that contains the instructions for the task.
    \item A rendered image of the intermediate sketch. In this image, the strokes of the input sketch will be rendered in blue and others will be in black.
    \item The intermediate sketch in SVG format.
\end{itemize}

\begin{figure}[!h]
  \centering
  \includegraphics[width=0.9\linewidth]{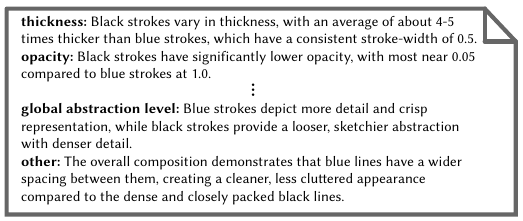}
  \caption{
\textbf{An example list of detected style differences.}
}
  \label{fig:detect_diff}
\end{figure}

% \todo{Should we put some exemplar differences...}
\asiahl{
In the preamble, we clearly outline the types of style differences we encourage the VLM to identify, including the global levels of abstraction and local stroke styles. 
For local stroke styles, we ask the VLM to focus on stroke thickness, smoothness, curvature, opacity.
\mbox{\autoref{fig:detect_diff}} shows an example list of detected style differences.
% geometric and appearance styles, such as the stroke width, smoothness, length, and opacity.
}
% \begin{minted}[fontsize=\footnotesize]{yaml}
% {
% "thickness": "Black strokes vary in thickness, with an average of about 4-5 times thicker than blue strokes, which have a consistent stroke-width of 0.5.",
% "opacity": "Black strokes have significantly lower opacity, with most near 0.05 compared to blue strokes at 1.0.",
% "global abstraction level": "Blue strokes depict more detail and crisp representation, while black strokes provide a looser, sketchier abstraction with denser detail"
% }
% \end{minted}

\paragraph{Adjustment code generation.}
In this step, we provide the VLM with the following information:
\begin{itemize}
    \item A preamble that contains the instructions for the task.
    \item A list of detected style differences.
    \item The intermediate sketch in SVG format..
    \item The augmented text prompt.
    \item A snippet of the skeleton style adjustment code.
    % that specifies how to read and write an SVG file, and defines a section for which the VLM should fill in the code for adjustment of the intermediate sketch. 
\end{itemize}
The VLM then completes the missing part of the skeleton code snippet, yielding a style adjustment code that specifies how to adjust the newly generated strokes to address the differences detected in the previous step.
For example, based on the detected differences in \autoref{fig:detect_diff}, the VLM generates the following adjustment code:
% \begin{minted}[fontsize=\footnotesize]{python}
% def adjust_stroke_style(path_data):
%     parsed_path = parse_path(path_data)
%     if "bold" in stylized_prompt_lower:
%          width_scale_factor *= 1.2   
%      for seg in parsed_path:
%         seg = seg.width * width_scale_factor
% \end{minted}
% \ichao{Change the following example and make it shorter..}
% Meanwhile, the VLM generate the following complex code to simplify the path structures of $\intermediatesketch$:
\begin{minted}[fontsize=\scriptsize]{python}
def adjust_stroke_style(path_data):
  for elem in parent.findall('.//svg:path', namespace):
    style = elem.attrib
    if 'stroke' in style and style['stroke'] != 'rgb(51, 102, 178)':
       # Get current attributes
       stroke_width = float(style.get('stroke-width', 2.0))
       stroke_opacity = float(style.get('stroke-opacity', 1.0))
       # 1. Remove extreme attributes
       if stroke_opacity < 0.12:
           parent.remove(elem)
           continue
       # 2. Adjust stroke width
       style['stroke-width'] = str(max(0.5, min(stroke_width, 
       given_sketch_style['stroke-width'] * 0.9)))
       # 3. Adjust opacity
       style['stroke-opacity'] = str(min(max(stroke_opacity, 0.95), 1.0))
       # 4. Adjust path to increase distance (shift position slightly)
       d = style.get('d', '')
       new_d = re.sub(
           r"([MLC])\s*(-?\d+\.?\d*)\s*(-?\d+\.?\d*)",
           lambda match: f"{match.group(1)} 
           {float(match.group(2)) + 2} {float(match.group(3)) + 2}", d)
       style['d'] = new_d
\end{minted}

% Finally, we execute $\mathcal{C}$ on $\intermediatesketch$ to obtain $\completesketch$.
Please see supplement for the details of the preamble we provided to the VLM and other adjustment codes generated by the VLM.
\begin{figure}[!t]
  \centering
  \includegraphics[width=\linewidth]{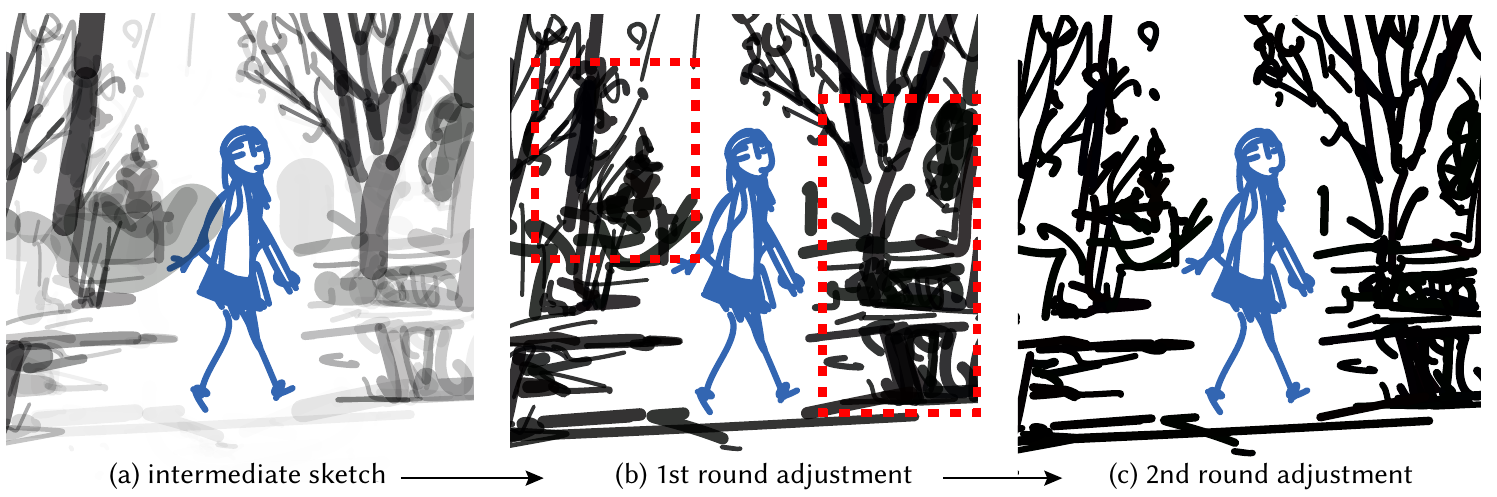}
  \caption{
  % \todo{Add some descriptions..}
\textbf{The iterative style adjustment process.}
Some style differences in the (a) intermediate sketch, like stroke width and opacity in red areas,  cannot be fully adjusted after (b) one iteration.  
% For example, the strokes width and opacity in the red areas.
These differences are addressed during (c) the second round of adjustment.
}
  \label{fig:iter_style_adjustment}
\end{figure}
We show an example of iterative style adjustment in~\autoref{fig:iter_style_adjustment}.

\section{Experiment}
\begin{figure*}[!t]
  \centering
  \includegraphics[width=\linewidth]{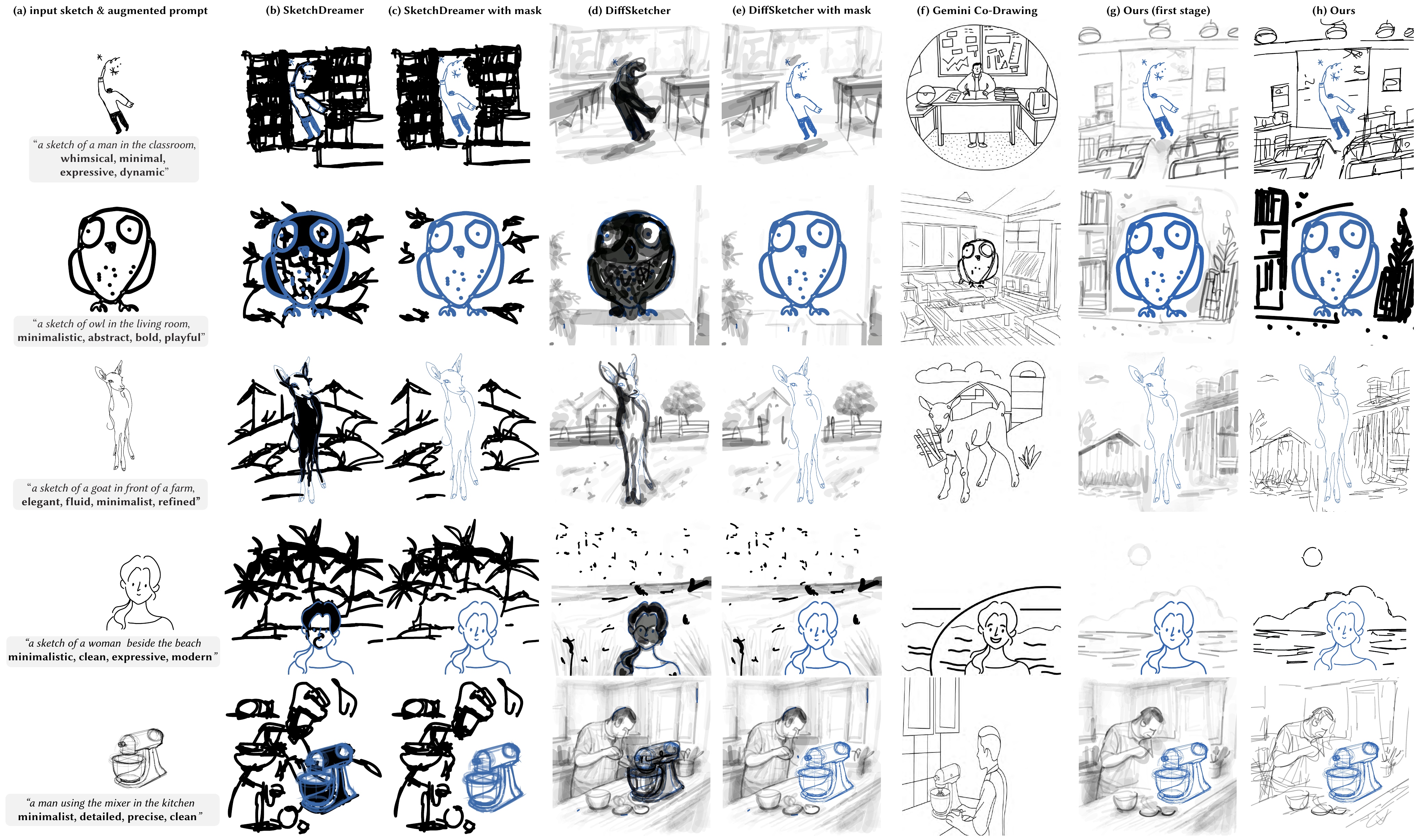}
  \caption{
  % \todo{I plan to change the layouts and content of this figure: remove vectorfusion result, add sketchdreamwer and diffsketcher results with mask exclusion...}
% \ichao{Check what is the prompt of these diffsketcher results used? Do they contain the stylized prompt?}
\textbf{Comparison with existing methods.} 
Given (a) the input sketch and the augmented prompt, \rvhl{(b, d) the results generated by SketchDreamer and DiffSketcher wrongly place too many strokes at the input sketch region and fail to match the styles of the input sketch.}
\rvhl{(c, e) We further remove the overlapping strokes using the input sketch mask to enhance the visual quality of their results.
However, these results still exhibit misaligned styles and often contains less desired content depicted in the input prompt.
}
% therefore alter the input sketch style. 
% (b) the results generated by VectorFusion cannot depict the desired content in the prompt.
% \asiahl{
% (c,d) SketchDreamer and DiffSketcher wrongly place too many strokes at the input sketch region and therefore alter the input sketch style. 
(e) Gemini Co-Drawing completes sketches that matches the prompt but fail to preserve the content and the style of the input partial sketch.
(f) The completed sketches generated by the first stage of our method avoids to place strokes overlap with input sketch, but the styles of strokes are still inconsistent.
% accurately represent the contents of the input prompts without altering the styles of the input sketch but the styles are inconsistent.
(g) Our full method further adjusts the styles of all strokes to match the styles of the input sketches.
% (\textbf{bold text}: style descriptions.)
}
  \label{fig:comparison}
\end{figure*}
\subsection{Implementation Details}
In this work, we use the GPT-4o model~\cite{hurst2024gpt} as the VLM, which extracts style descriptions and generates style adjustment codes.
We implement the first stage of our method using PyTorch~\cite{pytorch} and use the Adam~\cite{adam} optimizer to optimize the strokes.
\rvhl{We use $512$ strokes for stroke optimization for all cases.
During the style adjustment stage, some strokes are removed based on detected style differences, resulting in completed sketches that vary in density–some appearing more sparse while others are denser.
}
\rvhl{In terms of computation time}, it takes approximately $5$ minutes to optimize $1000$ iterations, while the second stage requires around $3$ minutes.
For all computations, we used a PC with an Intel i7CPU and an NVIDIA RTX 4080 GPU.

\asiahl{
The input sketches used in our experiment come from three sources: sketches generated using CLIPasso~\mbox{\cite{vinker2022clipasso}}, trace sketches found public website, and selected design sketches from the OpenSketch dataset~\mbox{\cite{gryaditskaya2019opensketch}}.
}
% (\eg~\mbox{\href{https://www.freepik.com/}{freepik}})

\subsection{Comparison with Existing Methods}
\label{sec:result_comparison}
% \todo{Maybe remove VectorFusion results because it is too bad. And we can move it to supplemental material...}
\asiahl{
We qualitatively and quantitatively compare our method to SDS-based sketch generation methods, including SketchDreamer~\mbox{\cite{qu2023sketchdreamer}} and DiffSketcher~\mbox{\cite{xing2023diffsketcher}}.
% Regarding VectorFusion, we use an public 3rd party implementation\footnote{\url{https://github.com/ximinng/VectorFusion-pytorch}} since the authors did not publicly shared official implementation.
For DiffSketcher, we replace the original vanilla Stable Diffusion using the ControlNet Scribble\footnote{\url{https://huggingface.co/lllyasviel/control_v11p_sd15_scribble}} used in SketchDreamer.
}
In~\autoref{fig:comparison}, we show the results generated by our method and all compared methods using identical user-provided partial sketch and stylized prompts.
\asiahl{
% The generated strokes using VectorFusion often do not match the desired content in the prompt. 
The results generated by SketchDreamer and DiffSketcher often place numerous strokes onto the input partial sketches instead of using them to illustrate desired content.
It reduce the amounts of strokes that can be used to depict desired content, and often results in missing content \rvhl{and low visual quality.}
\rvhl{We further reduce the overlapping strokes by using the input sketch mask.
However, eliminating these overlaps does not enhance the content being represented. 
More importantly, the styles of generated sketches do not match the style of the input sketches, resulting in inconsistencies in the final sketches.
}
In contrast, our method can generate strokes for depicting the desired content more effectively because we introduce the overlap penalty in~\mbox{\autoref{eq:complete_func}}.
}

\begin{figure*}[ht]
\begin{minipage}{0.58\textwidth}
  \includegraphics[width=\linewidth]{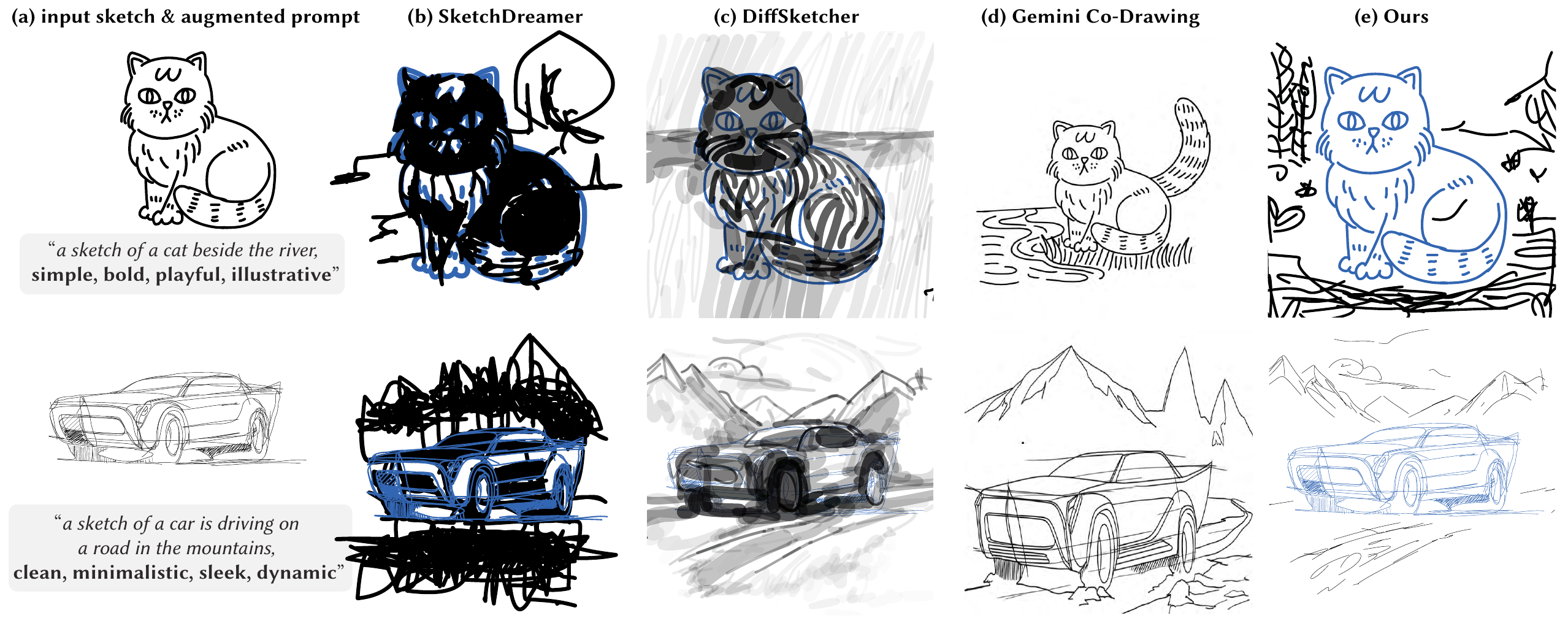}
  \caption{
\textbf{More comparison results.} 
Given (a) the input sketch and the augmented prompt, (b,c) the results generated by SketchDreamer and DiffSketcher wrongly place too many strokes at the input sketch region and therefore alter the input sketch style. 
(d) Gemini Co-Drawing completes sketches that matches the prompt but fail to preserve the content and the style of the input partial sketch.
(e) Our full method further adjusts the styles of all strokes to match the styles of the input sketches.
}
\label{fig:more_result}
\end{minipage}
\end{figure*}

\asiahl{
We further compared our method with Gemini Co-Drawing\footnote{\url{https://huggingface.co/spaces/Trudy/gemini-codrawing}} which is built on Gemini 2 native image generation.
As shown in~\mbox{\autoref{fig:comparison}}(e), although Gemini Co-Drawing can generate sketch that aligns with the input prompt, both the content and style of the input partial sketch are not well preserved.
}
% The partial sketches were prepared by re-tracing publicly shared sketches and clipart, or were generated by other sketch generation methods, such as CLIPasso~\cite{vinker2022clipasso}.
% The results of the ControlNet-based methods (\autoref{fig:comparison}(b,c)) often exhibit incomplete or inconsistent content.
% Additionally, these methods tend to apply style transformations that deviate significantly from those of the provided sketches, sometimes entirely altering the styles. 
In contrast, our method consistently completed sketches that faithfully represent the contents of the input text prompts with consistent styles.
% Also, the styles of the generated strokes and the provided sketches are consistent.
We show additional comparison results in~\autoref{fig:more_result}.

To further validate the effectiveness of our method in terms of preserving the sketch styles and completing the content, we gather an evaluation set containing $10$ sketches and perform two types of quantitative evaluation.
First, we use commonly use visual and text metrics to evaluate the performance of our method.
However, since these metrics are typically not used for evaluating the sketch completion task and have their own limitation, we additionally conduct an user evaluation which further validate our method.
\paragraph{Quantitative Evaluation using existing metrics.}
We used DreamSim~\cite{fu2023dreamsim} and DINO~\cite{caron2021emerging} as the visual metrics to measure the style consistencies and image similarities between the input partial sketches and the generated completed sketches.
Meanwhile, we assess the alignment between the content of each completed sketch and the input prompt using the VQA score~\cite{lin2024evaluating}.
The VQA score measures prompt-image alignment on compositional prompts more effectively than the CLIP score~\cite{radford2021learning} and is more closely aligned with human judgement.
\rvhl{It is important to note that the visual metrics we used, including DreamSim and DINO, measure both style and content similarity rather than focusing solely on style.
As a result, these metrics favor images that closely replicate the input sketch and leave the rest blank, which goes against our goal of completing the sketch based on the input prompt.
Therefore, it is essential to assess the results using both visual and text metrics.
}

As shown in~\autoref{tab:quan_result}, our method significantly outperforms the other methods across all visual metrics and achieves a comparable score on the text metric.
\asiahl{
We find that both the visual and text metrics alone used in this evaluation do not accurately reflect the preservation of the content and style from the input partial sketch, nor do they assess the quality of the completed sketch.
Regarding the visual metric, Gemini Co-Drawing achieves a high DINO score, but the input partial sketches are significantly altered.
On the text metric side, although both DiffSketcher and Gemini Co-Drawing receive higher scores, the completed sketch either lack quality or do not meet the preservation requirement necessary for our task.
}
% The evaluation set contains $10$ sketches.
% \unsure{We choose these visual metrics since there is no commonly used metrics to assess the style similarity between the partial sketch and a complete sketch.}
% However, visual metrics alone cannot be used to sufficiently evaluate performance because input partial sketches that do not receive additional strokes tend to achieve the best scores.
% Therefore, we also assess the alignment between the content of each completed sketch and the input prompt using the VQA score~\cite{lin2024evaluating} to eliminate the bias associated with visual metrics.
% The VQA score measures prompt-image alignment on compositional prompts more effectively than the CLIP score~\cite{radford2021learning} and is more closely aligned with human judgement.
% As shown in~\autoref{tab:quan_result}, our method significantly outperforms the other methods across all visual metrics and achieves a comparable score on the text metric.
% \unsure{Finally, although the quantitative results favor our method, we want to point out that both visual and text metrics used here are not commonly used for sketches.
% Moreover, the CLIP score used to assess the similarity between the completed sketches and input prompts are shown unreliable~\cite{sarto2025bridge}.
% Therefore, we provide the quantitative results here merely to provide an additional aspect for evaluation, and we would encourage readers to check the qualitative results and the user evaluation in~\autoref{sec:user_eval}.
% }
\begin{table}
\small
\centering
\begin{adjustbox}{width=\linewidth, center}
\begin{tabular}{c|cc|c}
\toprule
\multicolumn{1}{c}{} & \multicolumn{2}{c}{\textbf{Visual}} & \multicolumn{1}{c}{\textbf{Text}}\\
\cmidrule(lr){2-3} \cmidrule(lr){4-4}
 & DreamSim$\downarrow$ & DINO$\uparrow$ & VQA score $\uparrow$ \\
\midrule
% ControlNet LineArt &  0.415 & 0.545 & 0.463\\
% ControlNet Scribble &   0.592 & 0.365 & 0.709 \\
% VectorFusion & 0.541 & 0.451 & 0.426 \\
SketchDreamer  & 0.471 & 0.556 & 0.554 \\
DiffSketcher & 0.465 & 0.525 & \bestcell{0.816} \\
Gemini Co-Drawing & 0.365 & 0.584 & \seccell{0.804} \\
Our method &  \bestcell{0.290} & \bestcell{0.591} & 0.788 \\
\midrule
Our first stage & 0.434  & 0.488 & 0.332 \\
Our + Qwen3 & \seccell{0.305} & \seccell{0.578} & 0.781 \\
\bottomrule
\end{tabular}
\end{adjustbox}
\caption{
\textbf{Quantitative evaluation results.}
We compare our method to two SDS-based methods~\cite{qu2023sketchdreamer, xing2023diffsketcher} and Gemini Co-Drawing employing metrics that focus on visual and textual similarities.
Our method consistently outperforms the other methods for visual metrics and achieves comparable performance with other methods on textual metric.
We highlight the \besthint{first} and \sechint{second} best results.
% for each metric.
}
\label{tab:quan_result}
\end{table}
\paragraph{User evaluation.}
We conducted a user evaluation to further validate that our method generates sketches whose styles match those in user-provided partial sketches and depict complete content in the input prompt.
We use the same evaluation set used in~\autoref{sec:result_comparison} generated by our method, SketchDreamer and DiffSketcher.
\asiahl{
We chose to conduct user evaluations on these two methods without Gemini Co-Drawing because it often alters the input partial sketch significantly and only generates raster output.
}
Participants evaluated the quality of the generated completed sketches by conducting pairwise comparisons.
For each input sketch and prompt, we created three comparative pairs, ``Ours vs. SketchDreamer'' and ``Ours vs. DiffSketcher'', resulting in $20$ pairs for comparison.
During each comparison, two completed sketches were shown side by side in random order, along with their inputs.
Participants were asked to judge the sketches based on two criteria: ``How well they preserved the \textit{styles} of the input partial sketch'' and ``How effectively they depicted the \textit{content} of the input prompt''.
Each comparison was evaluated by $25$ different participants.
As shown in~\mbox{\autoref{tab:study_result}}, the participants preferred our method for both criteria.
\begin{table}
\small
\centering
\begin{adjustbox}{width=\linewidth, center}
\begin{tabular}{c|ccc|ccc}
\toprule
\multicolumn{1}{c}{} & \multicolumn{3}{c}{\textbf{Style}} & \multicolumn{3}{c}{\textbf{Content}}\\
\cmidrule(lr){2-4} \cmidrule(lr){5-7}
 & Ours & Others & neither & Ours & Others & neither \\
\midrule
% \sout{vs. LineArt} &  \bestcell{$98.4\%$} & $0.8\%$ & $0.8\%$ & \bestcell{$84.0\%$} & $9.6\%$ & $6.4\%$ \\
% % \midrule
% \sout{vs. Scribble} &  \bestcell{$96.8\%$} & $2.4\%$ & $0.8\%$ & \bestcell{$64.8\%$} & $30.4\%$ & $4.4\%$ \\
(a) vs. SketchDreamer & \bestcell{94.29} & 5.36 & 0.36 & \bestcell{86.43} & 3.21 & 10.36 \\
(b) vs. DiffSketcher & \bestcell{92.14} & 7.50 & 0.36 & \bestcell{55.71} & 40.36 & 3.93 \\
% vs. Gemini Co-Drawing & ? & ? & ? & ? & ? & ? \\
% Our method &  \bestcell{0.258}& \bestcell{0.270} & \bestcell{0.525} & 0.338 \\
\midrule
(c) vs. manual completion & \bestcell{76.79} & 23.21 & 0.0 & \bestcell{50.00} & 44.64 & 5.36 \\
\bottomrule
\end{tabular}
\end{adjustbox}
\caption{
% \ichao{TODO: update result with more participants.}
\textbf{User evaluation results.}
Compared to the two SDS-based methods and manual completion, the participants consistently preferred the completed sketches generated by our method in terms of both the style preservation and content depiction criteria.
(``Others'' denote to the comparing methods.)
We highlight the \besthint{best} result.
}
\label{tab:study_result}
\end{table}

\asiahl{
To further evaluate \mbox{\methodName}, we recruited two \rvhl{amateur} participants to manually complete two sketches and the results are shown in~\mbox{\autoref{fig:manual_comparison}}.
Participants were allowed to search for reference images online while completing the sketches.
We then conducted a user evaluation using these examples and show the results in~\mbox{\autoref{tab:study_result}}(c).
The user evaluation results indicate that our method obtained higher ratings for style preservation and content depiction compared to human participants.
Additionally, the participants noted that sketching multiple subjects interacting with one another while aligning the styles with the input sketch posed a particular challenge.
Therefore, they believe our method would be very helpful to allow users to concentrate more on sketching the content they can create.
}
% Additionally, we assessed the similarity between the content in the completed sketch and the input prompt using CLIP~\cite{radford2021learning}.
% The result in~\autoref{tab:quan_result} indicates that our method achieved similar scores to other ControlNet-based methods.
% This might be due to the result generated by ControlNet-based methods providing more details (\autoref{fig:comparison}), resulting in higher similarity scores with the input prompts.
% However, it is important to note that the sketch styles of those details often deviate significantly from the provided partial sketch.
% This indicates that our method successfully maintains the style consistency between a provided partial sketch and the generated strokes, and that the content is similar to those of other methods.

% \ichao{TODO: add quantitative evaluation, maybe using LPIPS or similar metric..}
% \paragraph{Quantitative evaluation}

\subsection{Diverse Sketch Scenario}
\begin{figure*}[ht]
  \begin{minipage}{0.48\textwidth}
  \includegraphics[width=\linewidth]{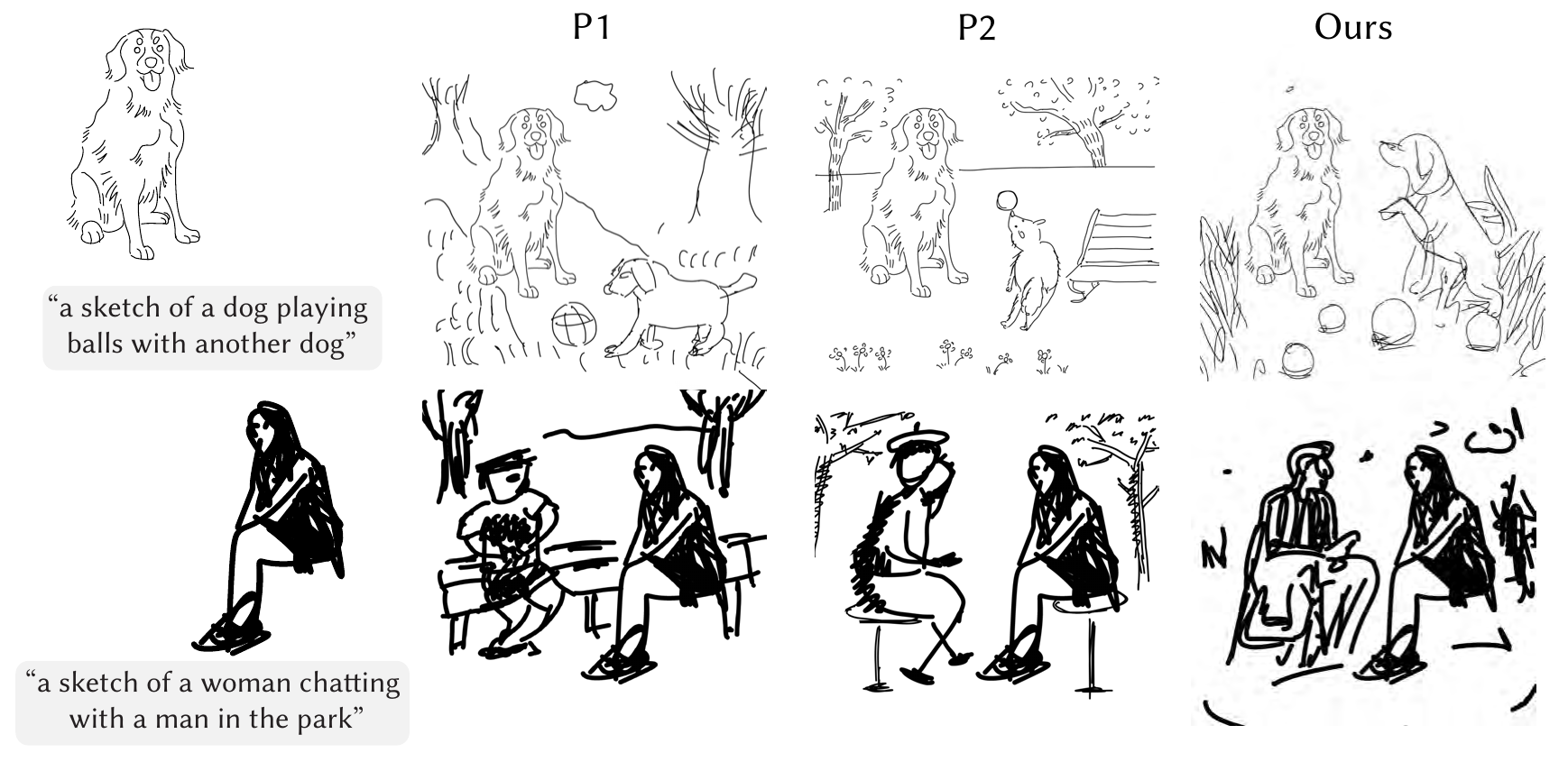}
  \caption{
\textbf{Comparison with manual sketch completion.} 
We recruited two amateurs to manually complete the input partial sketch based on the prompt.
Our method represents the subjects in the completed sketch more accurately and in a more cohesive style.
}
  \label{fig:manual_comparison}
      \end{minipage}
  \begin{minipage}{0.48\textwidth}
  \includegraphics[width=\linewidth]{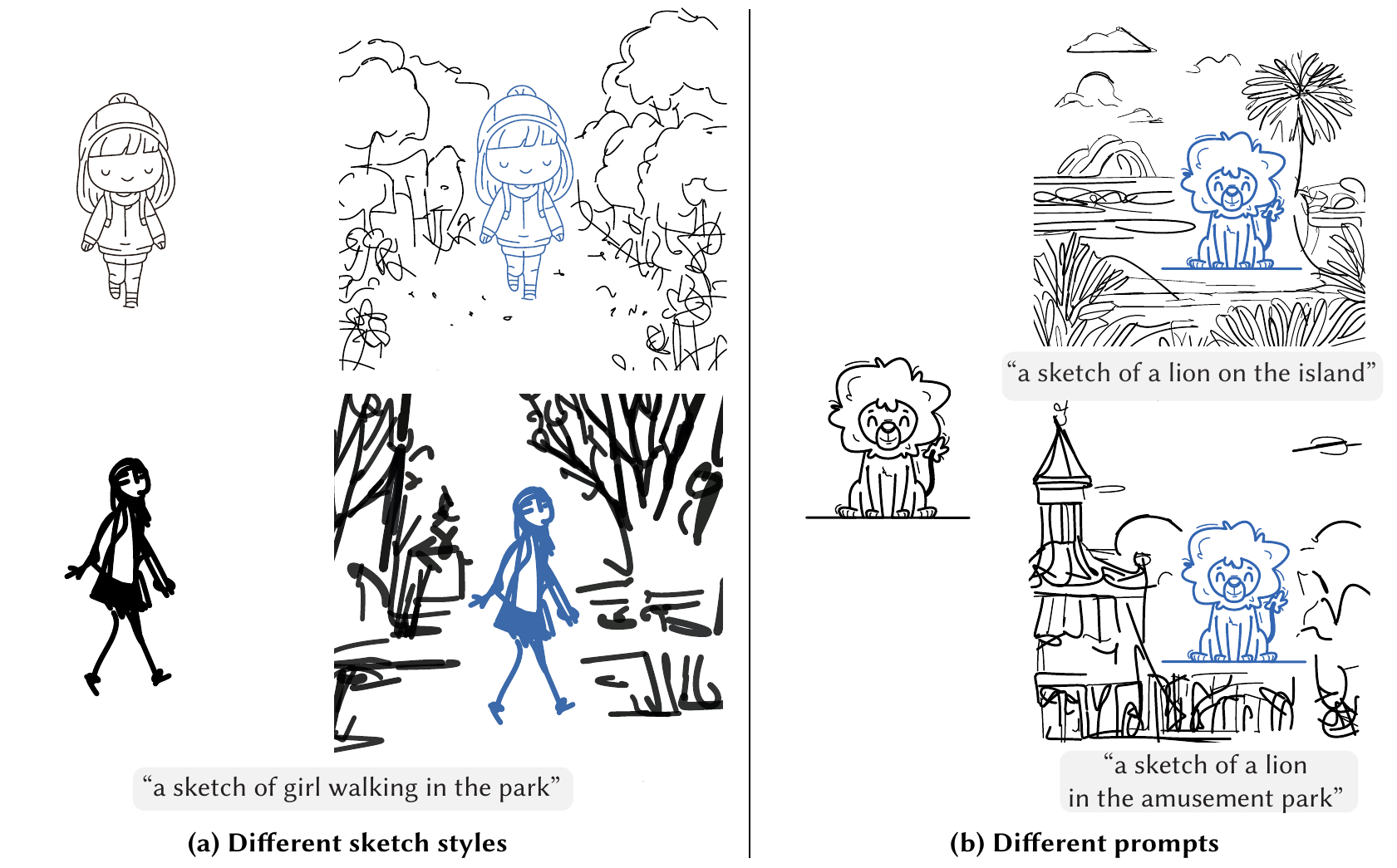}
  \caption{
\textbf{Various sketch scenarios.}
(a) Given the same prompt, our method can generate completed sketches that depict the same content in different styles that align with those of the user-provided partial sketches. 
(b) Given the same partial sketch, our method can generate different completed sketches representing the contents of various prompts.
}
  \label{fig:diff_promp_sketch}
      \end{minipage}
\end{figure*}

\begin{figure*}[!h]
  \centering
  \includegraphics[width=\linewidth]{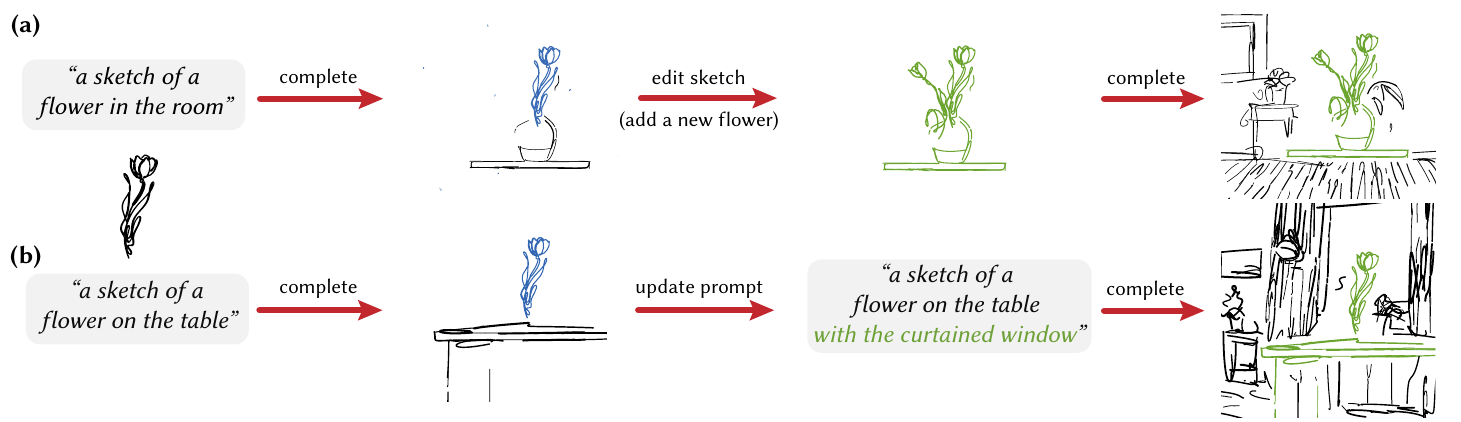}
  \caption{
\textbf{Examples of iterative sketch completion.}
% Given the same partial sketch, our method enables use
% \todo{Think about can we include a comparison with SketchDreamer here?}
After the initial sketch completion, the user can keep the strokes generated in the first completion and (a) edit the sketch or (b) update the input prompt .
Then, our method will complete the sketch once again to add more details.
(The blue and green line denotes the input partial sketch of the first and second iteration, respectively.)
}
\label{fig:iter_edit}
\end{figure*}

% In this section, we present the results obtained using various sketch scenarios to further demonstrate the utility of our method.
\paragraph{Iterative sketch completion.}
Sketching is often an iterative process, where users may want to introduce new details by adding new strokes or modifying the original prompt.
Our method enables users to achieve iterative sketch completion by retaining some strokes from the completed sketch and incorporating new ones (\autoref{fig:teaser}(b) and \autoref{fig:iter_edit}(b)), or by updating the input prompt (\autoref{fig:iter_edit}(a)).
\paragraph{Sketches with different prompts, or distinct sketches.}
Users may seek to employ a variety of partial sketches when generating sketches that depict the same content in the input prompt.
As shown in~\autoref{fig:diff_promp_sketch}(a), the completed sketches represent similar content but in different styles.
Additionally, as shown in~\autoref{fig:diff_promp_sketch}(b), the completed sketches created using different input prompts can represent distinct contents but share a similar style.

\subsection{Ablation Study}
\subsubsection{The effectiveness of the style adjustment stage} 
% \todo{Add quantitative results.}
We compared the results generated by only the first stage to those of our full method.
As shown in~\autoref{fig:comparison}(f), while the results of the first stage are both visually appealing and adequately represent the content of the input prompt, the sketch styles do not align well with those of the user-provided partial sketches.
In contrast, as shown in~\autoref{fig:comparison}(g) and\rvhl{~\mbox{\autoref{tab:quan_result}}}, our full method completes sketches that are better aligned with those of the user-provided sketch.
% \rvhl{We also show the quantitative results in~\mbox{\autoref{tab:quan_result}}.}
% , including aspects such as the stroke width, spacing, curvatures, are better aligned.
% Additionally, our full method achieved better quantitative results on the evaluation set across most of the metrics, as presented in~\autoref{tab:ablation_result}.

\subsubsection{Generalization of VLMs}
% \todo{Maybe move to supplemental material?}
% \input{fig_files/vlm_ablation_fig}
% \todo{Add both quantitative and qualitative evaluation results of open-source VLM here as well.}
Our method can utilize different VLMs to augment the input prompt and adjust style.
\rvhl{
In~\mbox{\autoref{fig:vlm_ablation}}, we show completed sketches using Gemini VLM~\mbox{\cite{team2023gemini}} and a open-sourced VLM Qwen3~\mbox{\cite{yang2025qwen3}}.
}
\rvhl{
Also, we present the quantitative results of using Qwen3 in~\mbox{\autoref{tab:quan_result}}.
These results demonstrate that our method, utilizing different VLMs, can achieve results comparable to our original method using GPT-4o.
% complete sketches in a consistent styles.
}

\subsubsection{The effectiveness of adaptive prompt augmentation}
\label{sec:abl_adap_prompt}
\begin{figure*}[!h]
  \centering
  % \begin{minipage}{0.68\textwidth}
  \includegraphics[width=\linewidth]{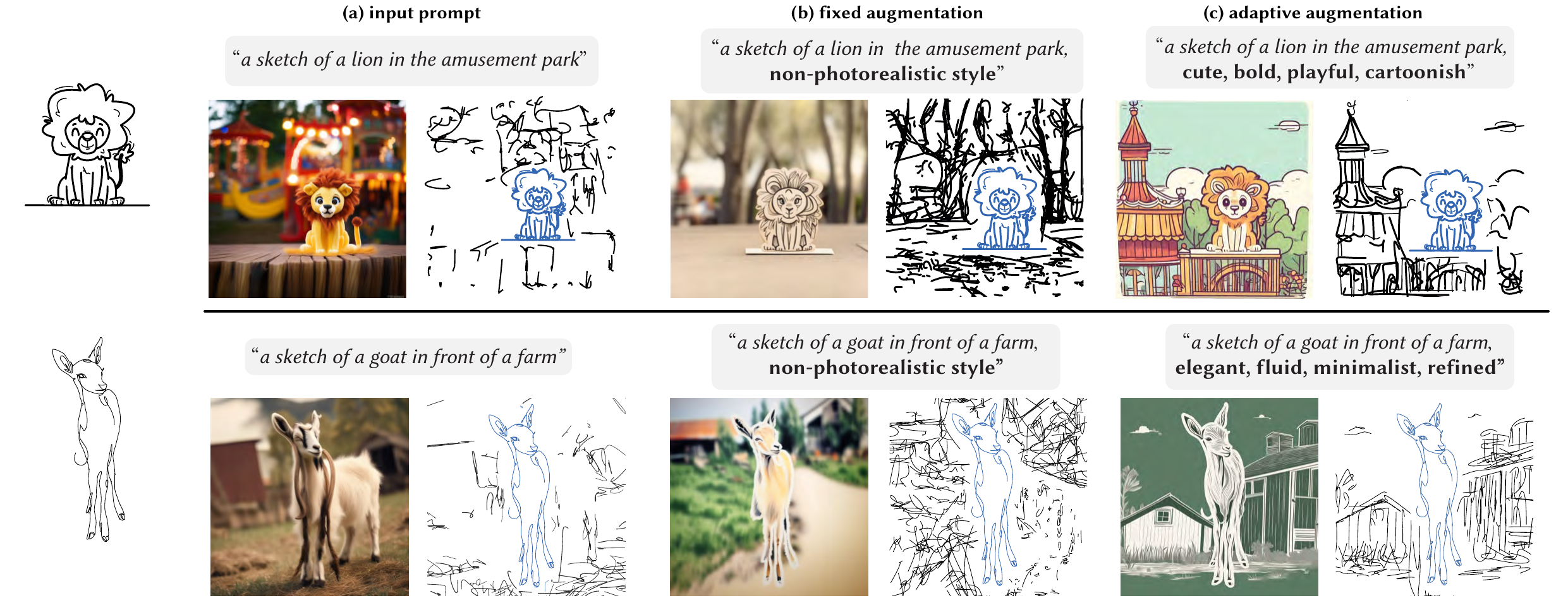}
  \caption{
\textbf{Prompt augmentation ablation study examples.}
(a,b) The guidance image generated using the partial sketch and the original input prompt or prompt with fixed augmentation contain unsuitable blurs and lack of clear boundaries.
% enough details for depicting the content in the prompt.
% align with the style of the partial sketch. 
Thus, our method could not then generate a final completed sketch that accurately depicted the complete content and the desired style.
(b) In contrast, the prompt with adaptive augmentation created a guidance image that allows our method to create a more completed sketch.
(The \textbf{bold text} in the prompt are style descriptions generated by the VLM.)
}
\label{fig:prompt_stylization_ablation}
% \end{minipage}
\end{figure*}

% \todo{Add fixed augmentation results.}
Our method enhances the input prompt by incorporating VLM-generated style descriptions of the input partial sketch, guiding the optimization of the intermediate sketch.
To assess its effectiveness, we compared results generated using the input prompt, our proposed adaptive augmented prompt, \asiahl{the input prompt, and a fixed augmented prompt that appended ``in non-photorealistic styles'' to each input prompt.}
\autoref{fig:prompt_stylization_ablation} shows that our adaptive augmented prompt generates outputs that better reflect the desired content and align with input sketch style.
Additionally, the quantitative results in \autoref{tab:ablation_result}(a,b) support these findings.
\rvhl{
We only report the text metric (VQA score), as the visual metrics tend to favor results that include only the input sketch with blank or minimal strokes, due to the content-style entanglement issue discussed in~\mbox{\autoref{sec:result_comparison}}.
}

\begin{table}
\small
\centering
\begin{adjustbox}{width=\linewidth, center}
\begin{tabular}{ccccc}
\toprule
% \multicolumn{1}{c}{} & \multicolumn{3}{c}{\textbf{Visual}} & \multicolumn{1}{c}{\textbf{Text}}\\
% \cmidrule(lr){2-4} \cmidrule(lr){5-5}
 & (a) w/o aug & (b) fixed aug & (c) VLM edit & (d) Ours \\
\midrule
% \midrule
VQA score $\uparrow$ & 0.443 & 0.532 &0.331 & \bestcell{0.788} \\
% \midrule
% DreamSim $\downarrow$ & ? & 0.367 & - & ? & 0.290\\
% DINO $\uparrow$ & ? & 0.560 & - & ? & 0.591\\
% b) w/o stylized prompt & ? & ? & ? & ?\\
% (c) w/o code generation & ? & ? & ? & ?\\
% \cdashline{1-5}[3pt/3pt]
% Our method &  \bestcell{0.258} & 0.270 & \bestcell{0.525} & \bestcell{0.338} \\
\bottomrule
\end{tabular}
\end{adjustbox}
\caption{
% \todo{Add Qwen3 result and revise captions.}
% \todo{Check visual metrics again.}
\textbf{Quantitative results of ablation study.}
Our method completes sketches with higher alignment with the input prompt compared to other alternatives.
% Meanwhile, (d) our method with another VLM Qwen3 performs comparably with our method with GPT-4o. 
% This suggests our method is generalized to other VLMs.
We highlight the \besthint{best} result for each metric.
}
\label{tab:ablation_result}
\end{table}

% VQA score $\uparrow$ & 0.578 & 0.532 &0.331 & ? & \bestcell{0.788} \\
% \input{fig_files/code_adjust_ablation_fig}
\subsubsection{The effectiveness of generating style adjustment code}
We found that requesting the VLM to generate style adjustment code leads to sketches that are more consistent in styles and contain more complete content than directly editing the SVG code.
As shown in~\autoref{fig:code_ablation}, some strokes may disappear if we request the VLM to adjust styles directly.
In contrast, the style adjustment code preserves the content when adjusting the styles to match those of the user-provided sketch.
Similar quantitative result is observed in~\autoref{tab:ablation_result}(c).
\rvhl{We also only report the results of VQA score because of the same reason as~\mbox{\autoref{sec:abl_adap_prompt}}}.

\section{Limitations and Future Work}
\begin{figure*}[ht]
  \begin{minipage}{0.48\textwidth}
  \includegraphics[width=\linewidth]{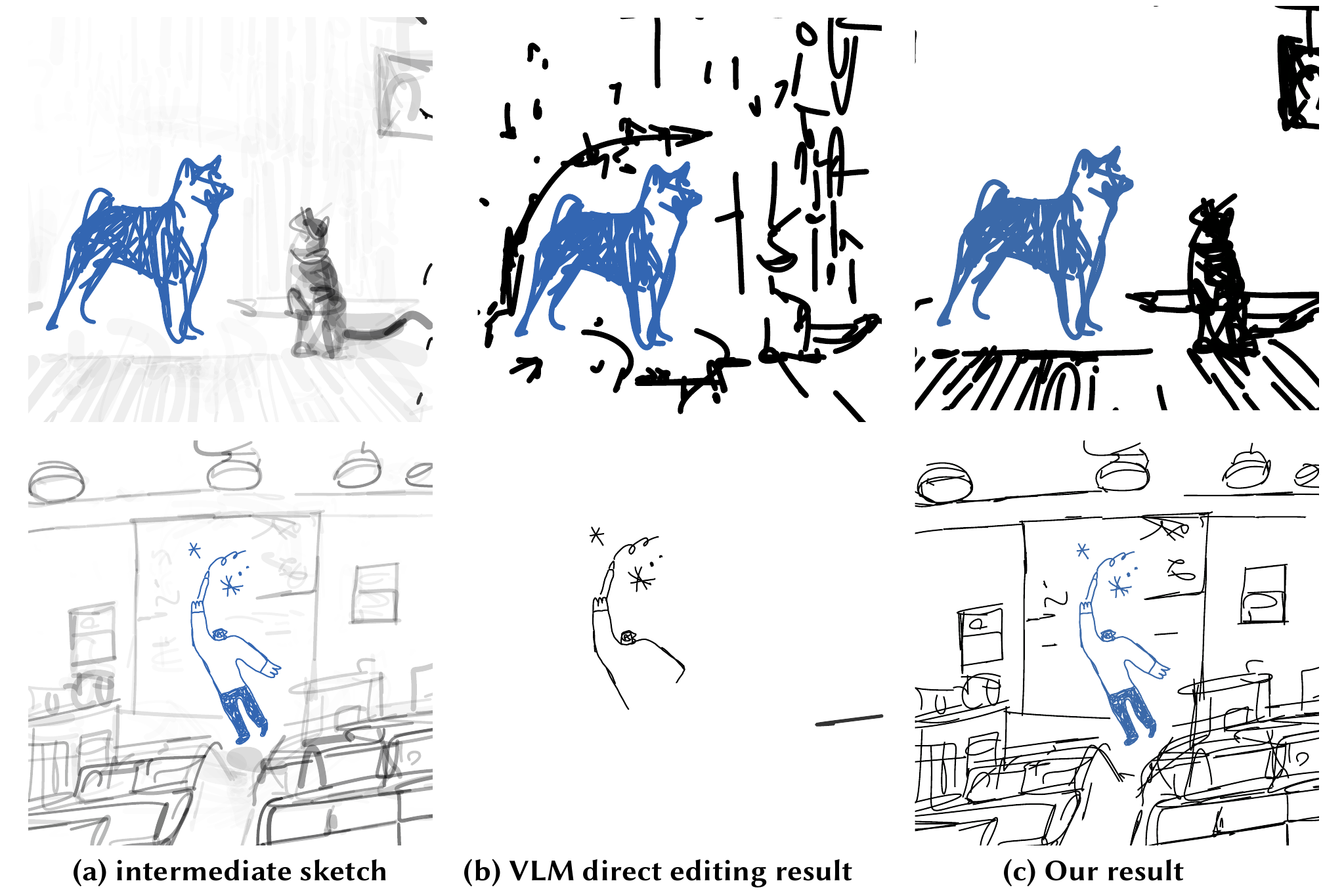}
  \caption{
  % \todo{Add another example.}
\textbf{Style adjustment code ablation study examples.}
Many important strokes depicted in (a) the intermediate sketch are missing from (b) the VLM direct editing results. 
In contrast, (c) our method effectively preserves such strokes while adjusting the styles. 
}
  \label{fig:code_ablation}
\end{minipage}
\begin{minipage}{0.48\textwidth}
  \includegraphics[width=\linewidth]{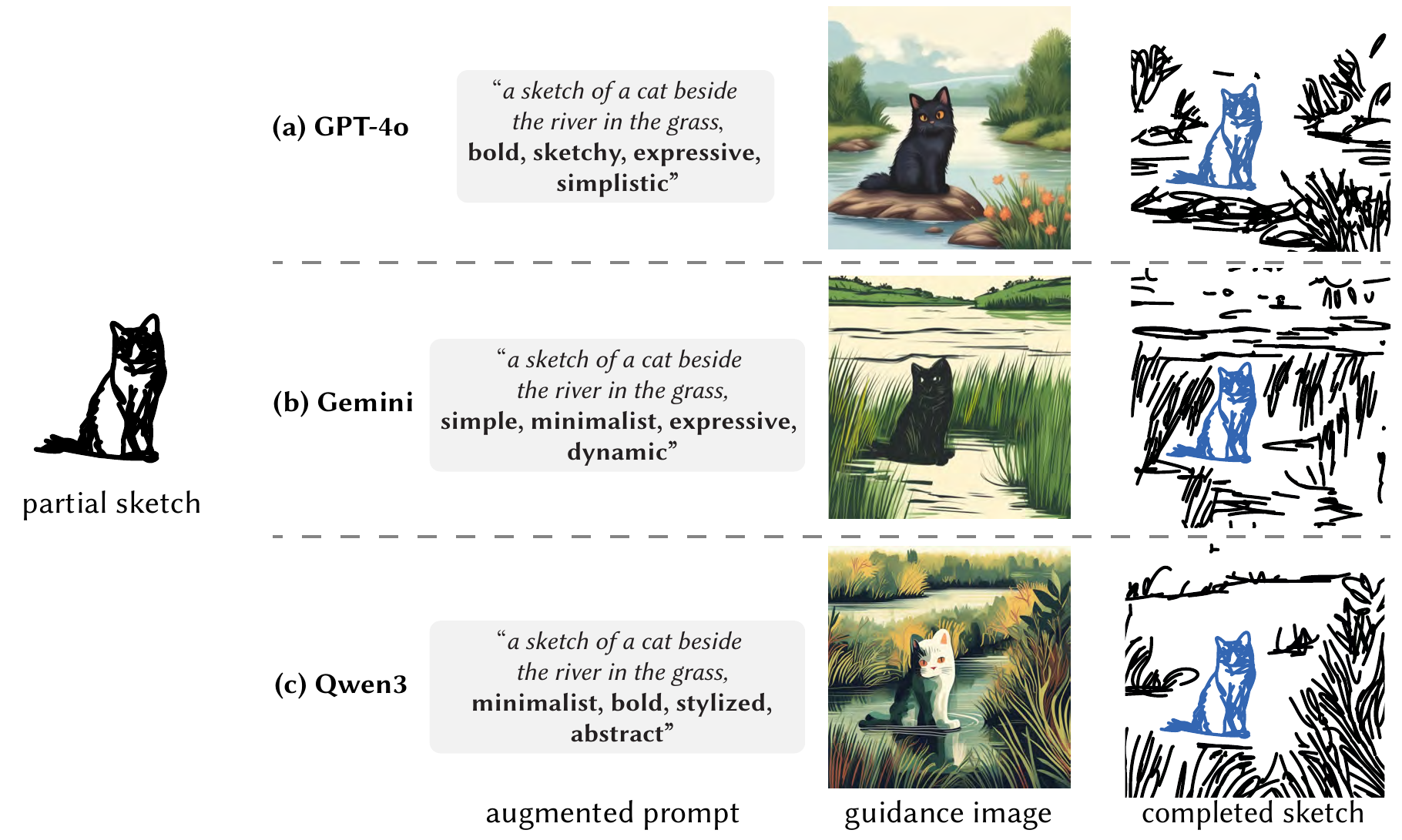}
  \caption{
  % \todo{replace and add one or two results of Qwen3.}
\textbf{VLM generalization example.} 
% \textbf{VLM generalization example.} 
Our method utilizing (a) GPT-4o, (b) Gemini, and (c) Qwen3 as the VLM can generate completed sketches that exhibit similar content and style based on the input partial sketch.
}
  \label{fig:vlm_ablation}
\end{minipage}
% \begin{minipage}{0.48\textwidth}
%   \includegraphics[width=\linewidth]{figs/vlm_ablation_v0.pdf}
%   % \includegraphics[width=\linewidth]{figs/VLM_adjust.pdf}
%   \caption{
% \textbf{VLM generalization example.} 
% % \textbf{VLM generalization example.} 
% Our method utilizing (a) GPT-4o and (b) Gemini as the VLM can generate completed sketches that exhibit similar content and style based on the input partial sketch.
% }
%   \label{fig:vlm_ablation}
% \end{minipage}
  \begin{minipage}{0.48\textwidth}
  \includegraphics[width=\linewidth]{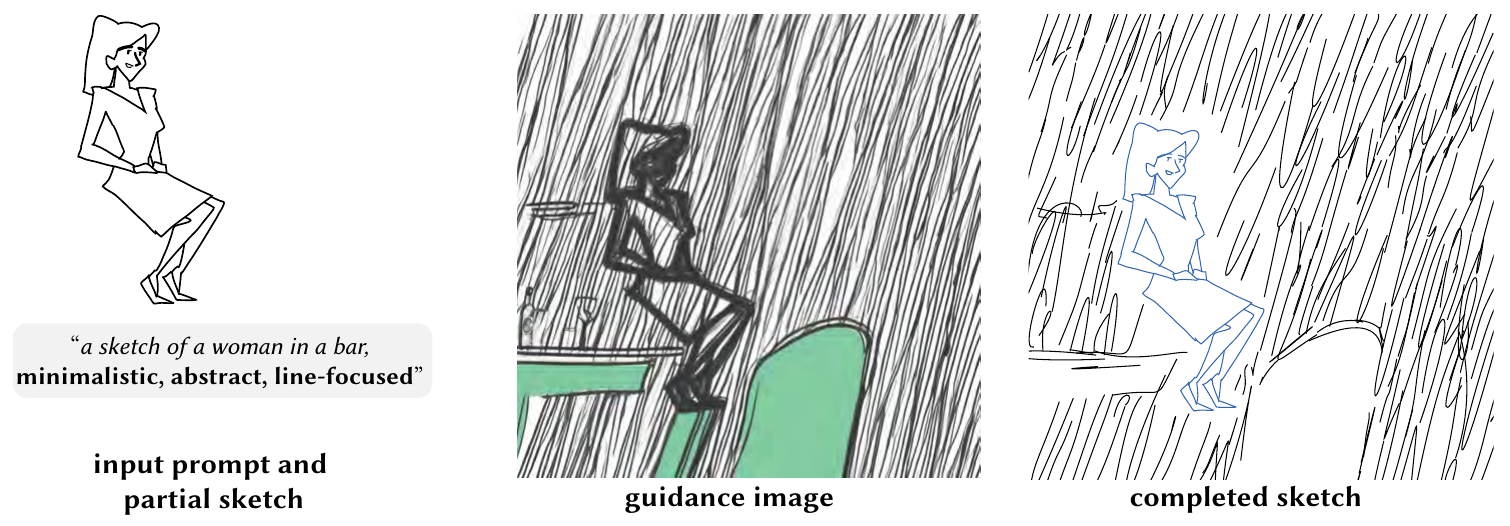}
  \caption{
  % \todo{Complete caption.}
\textbf{Limitation.} 
Our method cannot complete a sketch that accurately depicts the content of the input prompt and maintains the styles of the partial sketch with a broken guidance image generated by the ControlNet.
}
  \label{fig:limitation}
\end{minipage}
\end{figure*}

\paragraph{Reliance on large pretrained models.}
% \todo{revise here to mention that we would finetune the model in the future but need more data.}
Our method relies on two pretrained models: ControlNet for generating the guidance images and a VLM for augmenting the input prompt and create style adjustment codes.
Consequently, these models may occasionally generate unsatisfactory results.
As shown in~\autoref{fig:limitation}, if the guidance image lacks the specified content, our method cannot generate strokes that depict the desired content.
We will explore finetuning both ControlNet and the VLM to increase the robustness of our method.
% Even with clear guidance images, the VLM may not effectively adjust the strokes to represent the content while preserving the desired style (\autoref{fig:limitation}(b)).

\paragraph{Non-interactive generation.}
% Our method allows users to collaboratively create sketches using machine learning methods.
Our current method cannot generate completed sketch in real-time, which limits its application for interactive sketch completion.
We will explore \rvhl{possible speedup options, such as reducing optimization steps, lowering the number of stroke samples to minimize inside/outside checks (\mbox{\autoref{eq:inside_check}}), and running the VLM locally.}

% multi-scale stroke optimization to facilitate interactive sketch completion.

\paragraph{Does not support texture styles.}
Texture is a popular artistic styles, but our current adjustment method struggles to distinguish between strokes for geometry and those for texture.
We plan to classify these strokes and apply different style adjustment methods.

% During our current style adjustment phase, we have been splitting existing strokes into more segments to better support the desired style. 
% However, we have found this method seldom produces satisfactory results. 
% In the future, we plan to explore incorporating more strokes in the second stage to achieve a wider variety of styles.
\section{Conclusion}
In this paper, we introduce \methodName, a style-aware vector sketch completion method that accommodates diverse sketch styles by leveraging a pretrained VLM.
Our method allows users to input a sketch and automatically completes any missing content based on the specified prompt in a coherent style.
\rvhl{Additionally, users can adjust strokes in the input sketch, and our method will adjust the styles of the remaining strokes to match the adjustments.}
We demonstrate that the style descriptions extracted by the VLM from the input sketch enable our method to accurately complete the sketch regarding the desired content.
Furthermore, the style differences identified by the VLM between strokes enable our method to adjust the styles of strokes and achieve cohesive style.
Extensive experiment results indicate our method is effective across various sketch scenarios.

\bibliographystyle{ACM-Reference-Format}
\bibliography{paper}

\clearpage
% \begin{figure*}[!h]
%   \centering
%   \includegraphics[width=\linewidth]{figs/iter_edit_examples.pdf}
%   \caption{
%   % \todo{Finish the figure.}
% \textbf{Examples of iterative sketch completion.}
% % Given the same partial sketch, our method enables use
% After the initial sketch completion, the user can keep the strokes generated in the first completion and (a) update the input prompt or (b) edit the sketch.
% Then, our method will complete the sketch once again to add more details in the updated prompt.
% (The blue and green line denotes the input partial sketch of the first and second iteration, respectively.)
% }
% \label{fig:iter_edit}
% \end{figure*}

% \begin{figure*}[ht]
% \begin{minipage}{0.4\textwidth}
%   \includegraphics[width=\linewidth]{figs/limitation_v2.pdf}
%   \caption{
% \textbf{Limitation.} 
% Our method cannot generate a completed sketch that accurately depicts the content of the input prompt and maintains the styles of the partial sketch with a broken guidance image generated by the ControlNet.
% }
% \label{fig:limitation}
% \end{minipage}
% \end{figure*}

\end{document}